\documentclass{article}

\usepackage{arxiv}

\usepackage[utf8]{inputenc} 
\usepackage[T1]{fontenc}    
\usepackage{hyperref}       
\usepackage{url}            
\usepackage{booktabs}       
\usepackage{amsfonts}       
\usepackage{nicefrac}       
\usepackage{microtype}      
\usepackage{comment}

\usepackage{amsmath}
\usepackage{blkarray, bigstrut}
\usepackage{float}
\usepackage{tikz}

\usetikzlibrary{calc,angles,patterns,patterns.meta,decorations.pathmorphing}
\title{Joint Parameter Discovery and Generative Modeling of Dynamic Systems}

\author{
  Gregory Barber, ~Mulugeta A.~Haile\thanks{Corresponding author: \href{mailto:mulugeta.a.haile.civ@mail.mil}{mulugeta.a.haile.civ@mail.mil}}, ~and ~Tzikang Chen \\
  Vehicle Technology Directorate\\
  U.S. Army Research Laboratory\\
  Aberdeen Proving Ground, MD 21005\\

}

\begin{document}
\maketitle

\begin{abstract}
\noindent
Given an unknown dynamic system such as a coupled harmonic oscillator with $n$ springs and point masses. We are often interested in gaining insights into its physical parameters, i.e. stiffnesses and masses, by observing trajectories of motion. How do we achieve this from video frames or time-series data and without the knowledge of the dynamics model? We present a neural framework for estimating physical parameters in a manner consistent with the underlying physics. The neural framework uses a deep latent variable model to disentangle the system’s physical parameters from canonical coordinate observations. It then returns a Hamiltonian parameterization that generalizes well with respect to the discovered physical parameters. We tested our framework with simple harmonic oscillators, $n=1$, and noisy observations and show that it discovers the underlying system parameters and generalizes well with respect to these discovered parameters.  Our model also extrapolates the dynamics of the system beyond the training interval and outperforms a non-physically constrained baseline model. Our source code and datasets can be found at this URL: \url{https://github.com/gbarber94/ConSciNet}.
\end{abstract}


\section{Introduction}

Neural networks have produced strong results across a range of disciplines and have addressed problems in Image classification \cite{article_Lecun_Bengio_1, article_ImageNet, article_review_image_classification}, Natural language processing \cite{brown2020language, gardner2018allennlp_natural_language, Hirschberg261_natural_language}, Protein folding \cite{senior2020improved}, and so on. These successes are in part due to a neural network's ability to extract and learn complex features and relations from data. Often these learned features and relations are not easily accessible and are referred to as the model's ``black box''. This is problematic for modeling physical systems where we often have strong prior knowledge of the underlying physics, constraints, and relations a network must learn. Standard neural networks do not have this knowledge and must learn directly from data. In practice, even for simple physical systems, this proves to be a challenging task with networks often failing to generalize a system's behavior beyond the training interval. To address this issue neural architectures have been designed to explicitly embed physical constraints and symmetries \cite{greydanus2019,mattheakis2019,toth2020hamiltonian,raissi2019pinns, han2021adaptable} in the description of classical dynamical systems. 

Classical dynamical systems are described by Hamiltonian mechanics\cite{reichl2009modern_hamiltonian_1, PhysRevE.53.1890_hamiltonian_2, 1977RuMaS..32..177V_hamiltonian_3}. The principle of Hamiltonian mechanics was first incorporated into the design of neural networks nearly three decades ago~\cite{article_oldHNN}. In recent years this area has seen renewed interest \cite{greydanus2019, toth2020hamiltonian, Bertalan_2019,mattheakis2020hamiltonian}. Here we will focus our discussion around \cite{greydanus2019}'s Hamiltonian Neural Network architecture for learning exactly conserved quantities from data in an unsupervised manner.
Hamiltonian neural networks (HNNs) draw their inspiration from Hamiltonian mechanics and function through parameterizing a system's Hamiltonian function with a neural architecture. The Hamiltonian of a system $\cal{H}$ is a function that relates the system's total energy to its canonical coordinates i.e. its generalized coordinates $\mathbf{q}$ and momentum $\mathbf{p}$.
\vskip 0.1in
\noindent The Hamiltonian of a physical dynamical system is given by:

\begin{equation}
E_{total} = \cal{H}(\mathbf{q},\mathbf{p})
\label{Ham}
\end{equation}
\noindent
Where the total energy $E_{total}$ is the sum of the system's kinetic and potential energies. The dynamics or the time evolution of the states is given by:

\begin{equation}
\begin{split}
\frac{d \mathbf{q}}{dt} & = \phantom{-}\frac{\partial \cal{H}}{\partial \mathbf{p}} \\ \frac{d\mathbf{p}}{dt} & = - \frac{\partial \cal{H}}{\partial\mathbf{q}}
\end{split}
\label{Time_ev}
\end{equation}
\noindent
In a HNN, the model receives observations of a system's canonical coordinates as input. This input is passed through a feed-forward neural network and an energy-like value $H^*$ is output from the network. This section of the model is a neural network parameterization of Equation \ref{Ham}:
\begin{equation}
H^*=\text{HNN}(\mathbf{q},\mathbf{p})
\end{equation}
\noindent
The partial derivatives with respect to the input canonical coordinates of this network parameterization are then computed with the automatic differentiation library Autograd \cite{maclaurin2015autograd}. This section of the model is a parameterization of Equation \ref{Time_ev}:
\begin{equation}
\begin{split}
\hat{\frac{d \mathbf{q}}{{dt}}} & = \phantom{-}\frac{\partial H^*}{\partial \mathbf{p}} \\ \hat{\frac{d \mathbf{p}}{{dt}}} & = -\frac{\partial H^*}{\partial \mathbf{q}}
\end{split}
\label{time_ev_param}
\end{equation}
\noindent
The objective of a HNN model is to minimize the mean squared error metric between the networks partial derivatives, Equation \ref{time_ev_param}, and the derivatives of the input canonical coordinates $\frac{d \mathbf{q}}{dt}, \frac{d\mathbf{p}}{dt}$. After training, the network can be treated as the underlying time evolution model and can be evaluated over a time range with an ODE solver.

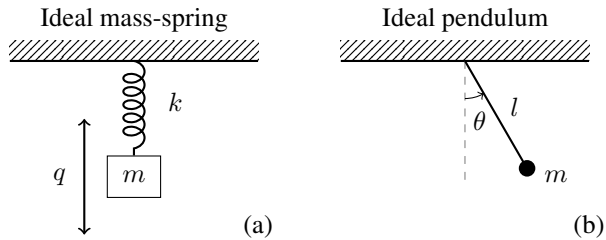
\begin{figure}[H]
\centering
\begin{tikzpicture}[scale=1.1]

\tikzstyle{mass} = [rectangle,fill=white,draw = black,inner sep=2mm]
\tikzstyle{spring} = [decoration={aspect=0.7, segment length= 5pt, amplitude= 4pt,coil},decorate, thick]

\node (a) [mass] at (0,0.6) {$m$};
\draw[decoration={aspect=0.7, segment length= 5pt, amplitude= 4pt,coil},decorate, thick] (0,2) -- (a);
\node[] at (.5,1.5) {$k$};

\fill [pattern = north east lines] (-1.5,2) rectangle (1.5,2.25);
\draw[thick] (-1.5,2) -- (1.5,2);
\coordinate (a1) at (-0.6,-.1);
\coordinate (a2) at (-0.6,1.3);
\draw[<->, thick] (a1) -- (a2);
\node[] at (-.9,0.6) {$q$};
\node[] at (0,2.5) { Ideal mass-spring};
\node[] at (1.5,0) {(a)};

\fill [pattern = north east lines] (2.5,2) rectangle (5.5,2.25);
\draw[thick] (2.5,2) -- (5.5,2);

\coordinate (origin) at (4,2);
\draw[dashed,gray,-] (origin) -- ++ (0,-1.5) node (orig_line) [black,below]{$ $};
\draw[thick] (origin) -- ++(300:1.5) coordinate (tip);
\fill (tip) circle (0.1);
\node[] at (4.6,1.45) {$l$};
\node[] at (5.1,.6) {$m$};
\node[] at (4,2.5) { Ideal pendulum};
\node[] at (5.5,0) {(b)};

\coordinate (a3) at (4.7,0.6);
\coordinate (a4) at (4,0.4);
\node[] at (4.18,1.28) {$\theta$};
\pic [draw, ->, angle eccentricity=1.5] {angle = orig_line--origin--tip};

\end{tikzpicture}
\caption{System diagrams. (a) Ideal mass-spring system, both the mass $m$ and the spring coefficient $k$ were varied during data generation. $q$ is the displacement from the neutral position. (b) Ideal pendulum system, the length parameter $l$ was varied during data generation and $m$ was held constant as it does not effect the period. $\theta$ is the angular position.}
\label{sys_diagram}
\end{figure}

This approach brings with it many aspects of the underlying Hamiltonian and is successful in generating predictions that respect the conservation of energy constraint outside the training interval. The HNN approach however does not maintain the same levels of flexibility and generative properties of the underlying Hamiltonian function in regards to a system's physical parameters. For example, let us consider the Hamiltonian for a simple pendulum system shown in Figure \ref{sys_diagram}.  

\begin{equation}
\mathcal{H} = mgl(1 - \cos(q)) + \frac{p^2}{2ml^2}
\label{pen_ham}
\end{equation}

Where $m$ is the mass, $g$ is gravitational constant, $l$ is length, $q$ is the position defined by $\theta$, $p$ is the angular momentum, $p = m\dot{\theta}l^2$. From this formulation, we can readily adjust the length parameter $l$ to simulate a new system. A HNN is incapable of this generalization. In \cite{greydanus2019} the system's physical parameters are held constant and absorbed into the canonical coordinates. In the current work we will draw inspiration from work in physical parameter discovery in particular SciNet \cite{iten2020}, to address this limitation and extend a HNNs generalization ability. 

In this paper, we will propose a neural framework for joint parameter discovery and generative modeling of Hamiltonian systems. We will then utilize this framework to disentangle the physical parameter of a Hamiltonian system from its canonical coordinates and return a Hamiltonian parameterization that maintains the flexibility and generative properties of the system's underlying Hamiltonian with regards to the system's physical parameters. We make the following contributions:
1. We propose a neural framework for join parameter discovery and generative modeling of Hamiltonian systems that merges deep latent variable models and physical constraint embedded neural networks, 2. We disentangle physical parameter like values from noisy coordinate observations, 3. We return a neural network parameterization of a system's Hamiltonian that generalizes well with regards to the discovered physical parameters and extrapolates the dynamics in physically consistent manner beyond the training interval. We will start with an overview of deep latent variables for parameter discovery with a focus on \cite{iten2020} and progress to our proposed architecture. 

\section{Parameter discovery}

Given an unknown dynamical system, we are often interested in gaining insight into its governing parameters. For simple linear systems, standard curve fitting and regression methods can return functions with interpretable parameters and coefficients. However, for high-dimensional nonlinear systems, this can prove challenging. Neural networks act as universal function approximators \cite{hornik1991approximation} and as such, they are a natural choice for modeling the function of such a system. These networks learn features and parameters from the data during training, to drive their predictive accuracy. The learned features and parameters are not easily interpreted in a physical context and are instead masked within the nonlinear combinations and transformations executed by the network. To acquire physically interpretable parameters we will turn to deep latent variable models, a family of neural architectures that utilize bottlenecks to extract minimal representations.

Deep latent variable models typically contain two components, an encoder that maps the input to a reduced latent space, and a decoder that utilizes the latent variables to make a prediction. The bottleneck encourages the encoder to learn a minimal representation of the input. A deep latent variable model's utility is limited by the quality of the learned latent representation. A good minimal representation will allow inference and the extraction of physically meaningful parameters. 

There are two major considerations when designing a deep latent variable model for parameter discovery: the size of the latent vector and latent variable disentanglement. The size of the latent vector is generally fixed prior to training. Here if we have prior knowledge of a system's minimal representation we can determine its size. For example, take an object moving with constant velocity, the minimal representation required to describe its next position includes its current position and velocity. The latent vector size for such a model could then be set to two. If we lack prior knowledge about the system the second consideration, disentanglement in the latent space becomes central. Ideally, a model with a good disentanglement will not encoded information into latent variables beyond those required for a minimal representation. For the case of an object with constant velocity, if the model was supplied with more than two latent variables (say six) it should only encode information in two. The SciNet architecture \cite{iten2020} has produced these types of disentangled representations for systems in classical and quantum mechanics.

The SciNet approach seeks to jointly disentangle a system's physical parameters from coordinate observations and model its behavior with a neural network parameterized function of the disentangled parameters and some auxiliary variables. SciNet's encoder is molded after that of a $\beta$-VAE \cite{Higgins2017betaVAELB} encoder, as input it receives coordinate observations $x$ and as output it returns  a reduced  set  of  latent  variables $Z$ : $f_{encoder}(x) = Z$. SciNet's decoder receives the latent variables and auxiliary variables as input and models a function of the input in terms of the target output: $f_{decoder}(Z,\ aux \ vars) = \hat{y}$. 

As a motivating example let us formulate a SciNet approach for an object moving with constant velocity. Given the objects initial position $q_0$ and velocity $v$, it's position at any time point $q(t)$ is given by: $q(t) = q_0 + vt$. SciNet aims to learn a parameterization of this function and its input parameters. To accomplish this we can vary the objects the speed and take an observations of its trajectory at each speed as input to the encoder, select $t$ as our auxiliary latent variable and seek to output $q(t)$ from the decoder. SciNet's decoder is then given by a function of $t$ and the latent variables output from the encoder:
$\hat{q(t)} = f_{decoder}(Z,t)$. 

As $t$ is provided to the system its minimal representation includes two constants: $q_0$, $v$. In this case, as $q_0$ is held constant it will not be disentangled by the network. During training SciNet's encoder will learn to map from an input trajectory length $n$ to a minimal latent representation encoding information into only one latent variable a velocity like a parameter: $v^*$. The decoder for this system can then be written as: $\hat{q(t)} = f_{decoder}(v^*,t)$. After training this parameterization can be split from the model and used to simulate new systems.  Similar to adjusting the velocity in the underlying equation, the velocity like parameter $v^*$ can be adjusted to simulate an object traveling a different speeds. This generative property is bounded by the range of the physical parameters present in the training data. 

The parameterization returned in a SciNet maintains a similar level of flexibility to a system's underlying function in regards to its physical parameters. SciNet however does not directly embed physical constraints and can return physically inconsistent predictions. In the current work, we will propose a novel HNN SciNet hybrid architecture for energy conserving systems. Our approach maintains the flexibility and generative properties of a system's underlying Hamiltonian in regards to a system's physical parameters. We call this framework of joint parameter discovery and generative modeling Constrained SciNet (ConSciNet).

\section{Constrained SciNet}

\begin{figure}[H]
    \includegraphics[width = 10.5cm, height = 4.5cm]{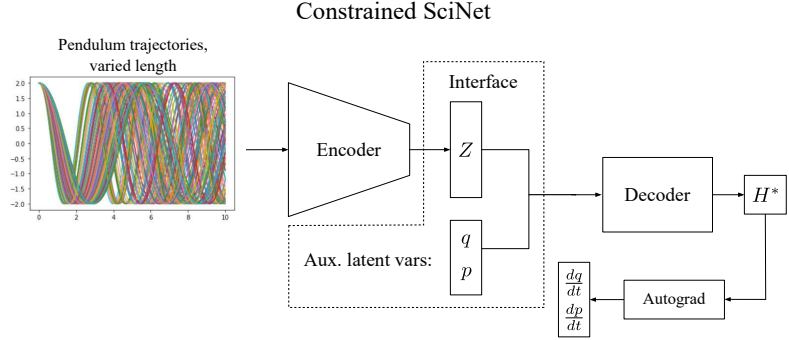}
    \centering
    \caption{ConSciNet architecture for a simple pendulum system. The Encoder architecture resembles that of a standard Variational AutoEncoder. It receives observations of a system's canonical coordinates as input and returns a reduced latent vector. The Interface concatenates this latent vector with a pair of canonical coordinates $(q,p)$ for the system. The Decoder is a modified HNN in addition to a canonical coordinate pair it takes in the latent vector. As in a HNN, the input is passed through a neural network to return a Hamiltonian-like value $H^*$. The partial derivatives of this network are then computed and mapped to the time evolutions, Equation \ref{time_ev_param}, and returned as the model's output.}
    \label{fig:arch}
\end{figure}

Constrained SciNet is a deep latent variable model that disentangles a system's physical parameters from canonical coordinate observations and returns a Hamiltonian parameterization that generalizes with respect to the discovered parameters.
Conceptually the proposed Constrained SciNet architecture follows a similar information flow to that of a standard SciNet with an important distinction in that the decoder has been replaced with a modified HNN and the auxiliary latent variables have been fixed by the system's canonical coordinates. The general layout of a ConSciNet model is illustrated in Figure \ref{fig:arch}. Structurally the ConSciNet model is composed of three components: an encoder, a decoder, and an interface. 

\vskip 0.1in
\noindent
\textbf{The Encoder} as in a standard SciNet is molded after the encoder of a Variational AutoEncoder \cite{kingma2013auto,rezende2014stochastic} with ELU activation functions \cite{clevert2015fast}. As input, the encoder receives observations of the system's canonical coordinates over some time interval. As output, the encoder returns a reduced latent vector $Z$. The size of $Z$ can be freely chosen prior to training. Here as we are interested in evaluating the model's performance in latent variable disentanglement we set the size of $Z$ to 3, a value greater than the minimum representations required for our trials. Our encoder is then given as: $f_{encoder}(observations) = (z_1,z_2,z_3)$.

\vskip 0.1in
\noindent
\textbf{The Interface}  receives the latent variables and auxiliary latent variables as input and structures this input for the decoder. Here the auxiliary latent variables are taken as one set of canonical coordinates per target. Limiting the coordinates passed to the decoder prevents the modified HNN decoder from directly learning a representation of the parameters itself and instead encourages the latent variables passed from the encoder to contain this representation. The Interface returns: $(q,p,z_1,z_2,z_3)$

\vskip 0.1in
\noindent
\textbf{The Decoder} is a modified HNN. In addition to the canonical coordinates it receives the latent variables as input: $H^* = \text{HNN}(q,p, z_1,z_2,z_3)$. As in a HNN the partial derivatives of this network are computed with respect to the canonical coordinates with Autograd \cite{maclaurin2015autograd}. The latent variables are treated as physical parameters and their partial derivatives are not returned. The inclusion of the physical parameter representation in the HNN's input allows the model to generalize with respect to a system's physical parameters, as in the underlying Hamiltonian formulation. Our decoder is given as: $f_{decoder}(q,p, z_1,z_2,z_3) = \frac{\partial H^*}{\partial p}, -\frac{\partial H^*}{\partial q}$.

\vskip 0.1in
\noindent
\textbf{The objective function} the ConSciNet model is trained with is given in Equation \ref{l_fn}:

\begin{equation}
\mathcal{L} = \Big(||\frac{\partial H^*}{\partial p} - \frac{dq}{dt}||_2 + ||\frac{\partial H^*}{\partial q} + \frac{dp}{dt}||_2 \Big) \: + \:  \Big( \beta \times D_{KL} \Big)
\label{l_fn}
\end{equation}

\noindent
This loss function is composed of two components. The first corresponds to the HNN loss and measures the performance of the decoder. The second component corresponds to the $\beta$-VAE loss (KL-divergence) and encourages a disentangled representation. Heuristically, we found small values for $\beta$ around 0.005 struck a good balance between latent variable disentanglement and model performance.    

\begin{figure}[H]
    \centering
    \includegraphics{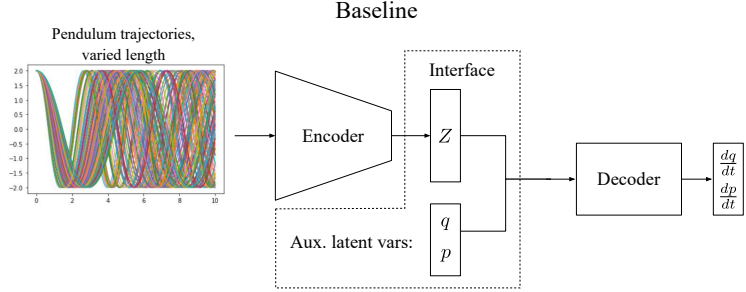}
    \caption{Baseline architecture for a simple pendulum system. The Encoder architecture resembles that of a standard Variational AutoEncoder. It receives observations of a system's canonical coordinates $(q,p)$ as input and returns a reduced latent vector. The Interface concatenates this latent vector with a pair of canonical coordinates for the system. The Decoder is a standard feed-forward network that returns the time evolutions}
    \label{fig:base_arch}
\end{figure}

\noindent
\textbf{Baseline model.}
In addition to the ConSciNet model, a Baseline model without the Hamiltonian parameterization was trained as a point of comparison. The architecture for the Baseline model is sketched in Figure \ref{fig:base_arch} and resembles that of a standard SciNet. The Baseline model's decoder did not parameterize the Hamiltonian and instead consisted of a standard feed-forward network that directly output $\hat{\frac{dq}{dt}},\hat{\frac{dp}{dt}}$. The objective function the Baseline model is trained with is given in Equation \ref{b_l_fn}:

\begin{equation}
\mathcal{L} = \Big(||\hat{\frac{dq}{dt}} - \frac{dq}{dt}||_2 + ||\hat{\frac{dp}{dt}} - \frac{dp}{dt}||_2 \Big) \: + \:  \Big( \beta \times D_{KL} \Big)
\label{b_l_fn}
\end{equation}

\vskip 0.1in
\noindent
\textbf{Implementation}. We implemented the ConSciNet and Baseline model in PyTorch \cite{paszke2019pytorch}. Our models and source code are available at \url{https://github.com/gbarber94/ConSciNet}.

\section{Trials}

The ConSciNet model's performance and parameter generalization ability was assessed on two simple dynamic systems from \cite{greydanus2019}: an ideal pendulum and ideal mass-spring system. For both systems, their physical parameters were varied during data generation and noise was added.

\subsection{Task 1: Ideal pendulum}

\noindent
The Hamiltonian of an Ideal pendulum system given in Equation \ref{pen_ham} was used to simulate multiple systems. For each simulation the length parameter was varied between 0.3 and 0.8 and its time evolution was evaluated over the range 0 to 10. For each time evolution, a noise vector of equal length was then added. The noise vector was sampled from a normal distribution with mean $\mu = 0$ and standard deviation $\sigma = 0.03$. In total 50,000 trajectories of length 100 were generated. These trajectories were used as input into the encoder. For every trajectory, a time point was randomly selected over the generation range. At this point, the system's time evolution was evaluated and its derivative stored. The coordinate evaluation at this point was used as the auxiliary latent variables and its derivative was used as the model's target.

\subsection{Task 2: Ideal mass-spring}

\begin{equation}
\mathcal{H} = \frac{1}{2}kq^2 + \frac{p^2}{2m}
\label{spring_ham}
\end{equation}

\noindent
The Hamiltonian of an Ideal mass-spring system is given in Equation \ref{spring_ham} where $k$ is the spring constant, $m$ is the mass, $q$ is the position and $p$ is the linear momentum given by $p = mv$. Here we sampled values for $k$ between 0.1 and 0.5, and values for  $m$ between 0.5 and 1. For each parameter pairs we simulated, the system's time evolutions from [0,10] and added a noise sampled from a normal distribution with mean $\mu = 0$ and standard deviation $\sigma = 0.03$. In total 50,000 trajectories of length 100 were generated. The model's targets and auxiliary latent variables were sampled following the same procedure described in Task 1.

\section{Results}

The performance of the ConSciNet and Baseline model (SciNet with a standard feed-forward decoder) were evaluated in terms of their latent variable disentanglement and generative functionality. After training, both models were split at their interface. In both tasks, the encoder was used to assess latent variable disentanglement. Trajectories with known physical parameters were passed to the encoders and their output latent variables were stored. The known physical parameters and their corresponding latent variables were then examined to identify latent variables encoding physical parameter like values.

Similarly, for both tasks, the decoder was used to assess the generative functionality of the models. The latent variable or variables identified as corresponding to physical parameters were interpolated between the minimum and maximum values of the corresponding physical parameters. The remaining latent variables that did not encode information were taken as 0. These values were then formatted to match the input latent vector structure of the decoder to create a parameter argument. For example, if the first entry in the latent vector corresponds to a length-like parameter $l^*$ and all others encoded no information the parameter argument would be given by: $[l^*,0,0]$. A parameter argument was generated for each interpolated latent variable. An ODE solver was then used to integrate the decoders and return coordinate predictions for each parameter argument. Here we used SciPy's \cite{2020SciPy-NMeth} \texttt{solve\_ivp} function to integrate the network with a tolerance of 1e-12. A 4th order Runge-Kutta \cite{runge1895numerische} method was selected for the solver. During this evaluation, an initial canonical coordinate value $(q = 1, p = 0)$ was supplied and the parameter argument was held constant. These network-generated trajectories were then assessed for physical constancy and alignment to the ground truth.  

\subsection{Ideal pendulum}  

\begin{figure}[H]
    \includegraphics[width = 10cm, height = 10.25cm]{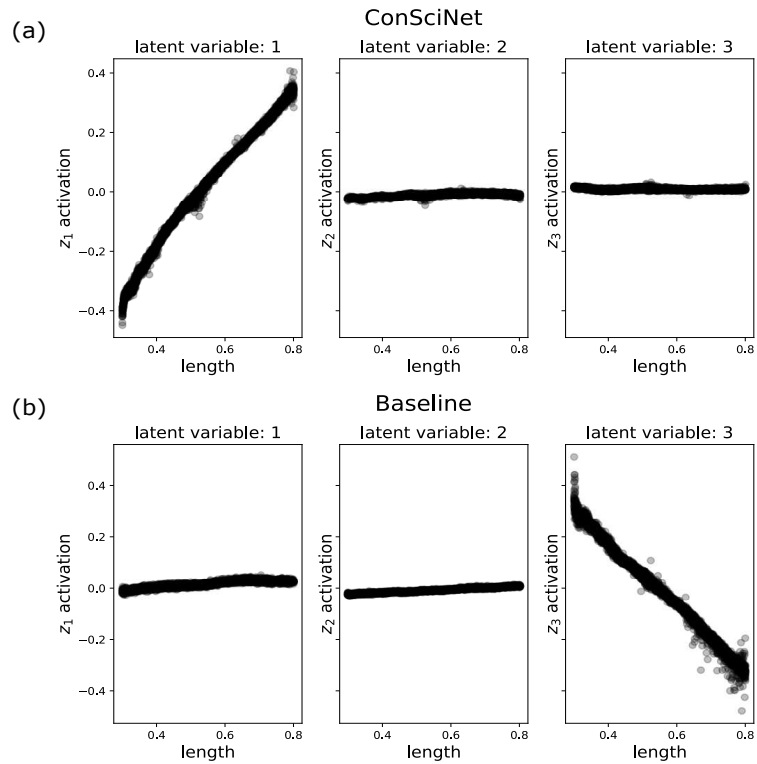}
    \centering
    \caption{Pendulum latent variable disentanglement. The y-axis corresponds to the values of the latent variables output from the Encoder. The x-axis corresponds to the ground truth length value for the input trajectories. In (a) the ConSciNet disentanglement is given. The ConSciNet model encoded a length like parameter in the first latent variable. In (b) the Baseline disentanglement is given. The Baseline model encoded a length like parameter in the third latent variable. The remaining latent variables in both models encoded little to no information and returned values close to zero.}
    \label{fig:penlv_dis}
\end{figure}

We found that both the ConSciNet and Baseline models were able to successfully disentangle a length-like parameter $l^*$ from the noisy input coordinates. In Figure \ref{fig:penlv_dis} we present the disentanglement results for both models. Here we plot the latent variable activations output by the encoder against the length value of the input trajectory. 

From Figure \ref{fig:penlv_dis}(a), the ConSciNet model encoded $l^*$ in the 1st latent variable and encoded values close to 0 in the remainder. From Figure \ref{fig:penlv_dis}(b) the Baseline model encoded the $l^*$ value in the 3rd latent variable and values close to 0 in the remainder. The specific position of the latent variable encoding information and its sign do not convey information. These values are subject to change if the model is retrained. The discovered parameter $l^*$ is not the exact underlying length parameter rather it is some transformation of the value. In this case, the transformation is roughly linear. 

\begin{table}[H]
\caption{MSE between the ground truth trajectory and model predicted trajectory for a given length parameter $l$. Each row corresponds to a column in Figure \ref{fig:gen_pen_trj}(a), where $l$ is mapped to $l^*$ and evaluated.}  
\centering
\begin{tabular}{lrrr}
\multicolumn{3}{c}{Pendulum interpolated | MSE: t-span[0,20]} \\
\toprule
  length &  Baseline &  ConSciNet \\
\midrule
     0.3 &      0.160261 &       0.028775 \\
     0.5 &      0.023244 &       0.000613 \\
     0.6 &      0.022039 &       0.000117 \\
     0.8 &      0.025094 &       0.001503 \\
\bottomrule
\end{tabular}
\label{table:mse_pen}
\end{table}

\begin{figure}[H]
    \includegraphics[width = 14cm, height = 6.5cm]{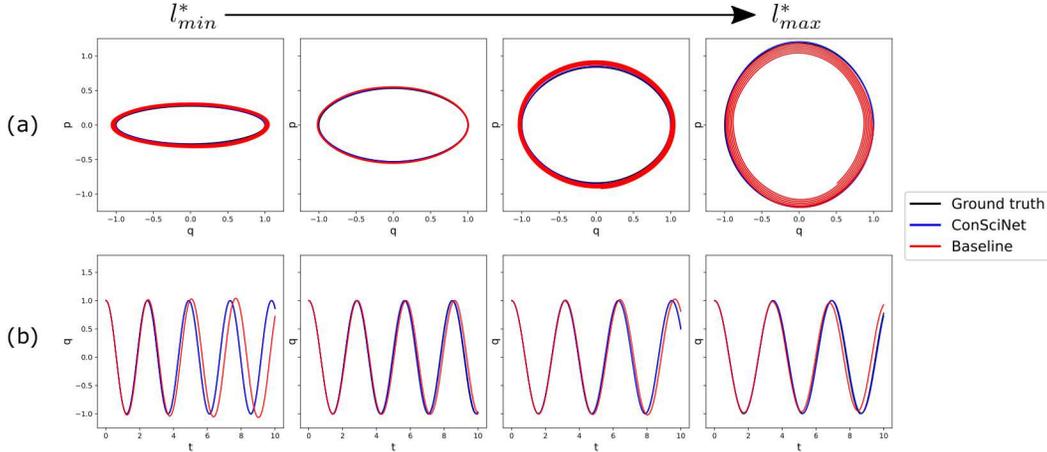}
    \centering
    \caption{Generating pendulum trajectories. The discovered length like parameter $l^*$ for each model was interpolated between its minimum and maximum values and used to evaluate each model's decoder. From left to right the value of $l^*$ input into each model's decoder was increased. (a) Phase plot of both models predictions and the ground truth values. The models are trained in the time interval [0,10] and predictions extrapolated over the interval [0,20]. (b) Time plot of the position $q$ over the training interval [0,10]. The ConSciNet model returned predictions inline with ground truth values and respected the conservation of energy constraint while the Baseline model failed to respect the constraint.}
    \label{fig:gen_pen_trj}
\end{figure}

We found the network parameterization learned in ConSciNet's decoder maintained similar generative properties to the system's underlying time evolution equations with regards to the physical parameters. ConSciNet was able to generalize across the discovered parameter domain and accurately extrapolate the time evolutions of the system governed by these parameters beyond the training interval t-span [0,10]. In Figure \ref{fig:gen_pen_trj} we present trajectories generated from evaluating the decoder at four values of $l^*$. 

The evaluation values for $l^*$ were selected using a scheme that allowed for a ground truth comparison. From Figure \ref{fig:penlv_dis}, it is apparent that the mapping between the ground truth parameter and the parameter-like value learned by the network can be modeled with a simple polynomial fit. Here we fit a cubic polynomial for each model mapping from $l$ to $l^*$. We then linearly interpolated four values for $l$ between the minimum and maximum values present in the data and mapped these values to their corresponding $l^*$ values to obtain the evaluation values of $l^*$ for each model.

In Figure \ref{fig:gen_pen_trj}(a), we present phase plots generated from evaluating the models at each $l^*$ value over the time interval t-span of [0,20]. The ConSciNet model generated trajectories that aligned with the ground truth trajectories and successfully respected the conservation of energy constraint over the evaluation window. The Baseline model on the other hand deviated from the ground truth trajectories over time and failed to conserve energy outside of the training interval, t-span of [0,10]. Its predictions generally gained or lost energy as the evaluation interval expanded beyond the training interval.

In Figure \ref{fig:gen_pen_trj}(b), we highlight the generalization ability of the models in regards to the discovered parameters, over the time interval t-span of [0,10]. As the length like parameter $l^*$ is increased the period decreases aligning with the expected physical behavior. The Baseline model while matching this behavior suffers in accuracy in comparison to the ConSciNet model. 

The individual results here for each value of $l^*$ are similar to \cite{greydanus2019}'s results for a system with fixed parameters suggesting the ConSciNet approach successfully generalizes an HNN with regards to a system's parameters. 

The success of the Baseline model in parameter disentanglement and it's predictive performance over the training range indicates that the SciNet framework can be used as in our approach to model time derivatives. This may have applications in parameter disentanglement for NeuralODEs. 

\subsection{Ideal mass-spring}

\begin{figure}[H]
    \includegraphics[width = 10.5cm, height = 8cm]{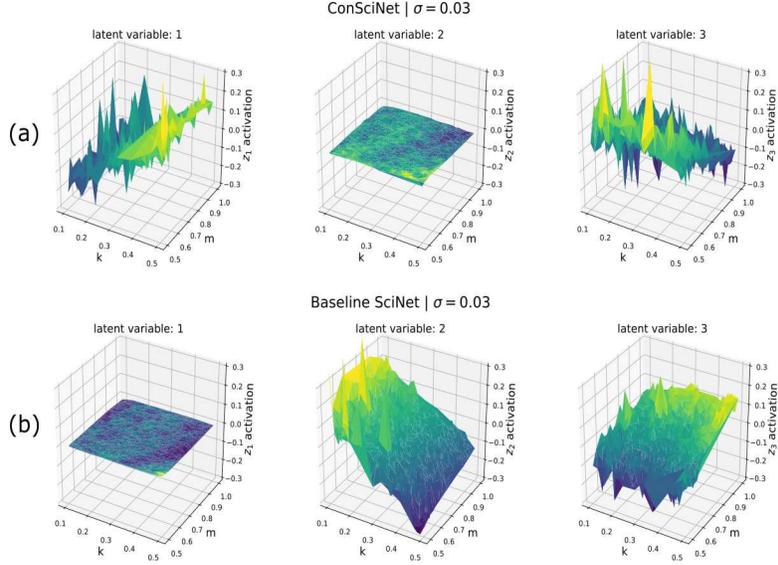}
    \centering
    \caption{Mass-spring latent variable disentanglement. The z-axis corresponds to the latent variables output from the Encoder. The x-axis corresponds to spring coefficient of the input trajectory. The y-axis corresponds to the mass value of the input trajectory. In (a) the ConSciNet model encoded a spring coefficient like value $k^*$ in the 1st latent variable a mass like value $m^*$ in the third latent variable and little to no information in the 2nd latent variable. In (b) the Baseline model encoded a spring coefficient like value $k^*$ in the 2nd latent variable, a mass like value $m^*$ in the third latent variable and little to no information in the 1st latent variable.}
    \label{fig:spring_lv_dis}
\end{figure}

\begin{figure}[H]
    \includegraphics[width = 13.5cm, height = 5.5cm]{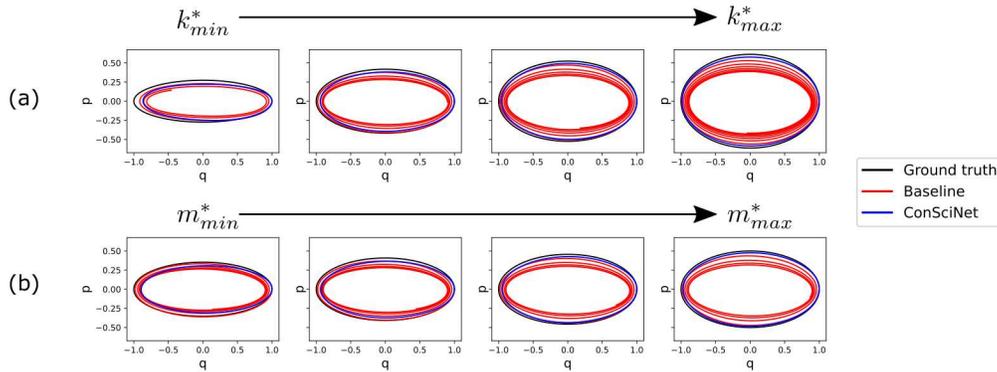}
    \centering
    \caption{ Generating mass-spring trajectories. The discovered spring coefficient like value $k^*$ and mass like parameter $m^*$ were interpolated between their minimum and maximum values and used to evaluate each model's decoder. In (a) $m^*$ was held constant and the evaluation value of $k^*$ was increased from left to right. In (b) $k^*$ was held constant and the evaluation value for $m^*$ was increased from left to right. Both (a) and (b) were evaluated over the extrapolated time interval [0,20] based on models trained over the interval [0,10]. The ConSciNet model respected the conservation of energy constraint while the Baseline model lost energy over time.}
    \label{fig:spring_trj}
\end{figure}

\begin{table}[H]
\centering
\caption{MSE between the ground truth trajectory and model predicted trajectory for a given spring coefficient $k$. Each row corresponds to a column in Figure \ref{fig:spring_trj}(a), where $k$ is mapped to $k^*$ and evaluated while $m^*$ is held constant at the value corresponding to $m = 0.75$.} 
\begin{tabular}{lrrr}
\multicolumn{3}{c}{Mass-spring interpolated $k$ | MSE: t-span[0,20]} \\
\toprule
  $k$ &  Baseline MSE &  ConSciNet MSE \\
\midrule
                0.1 &      0.185356 &       0.066921 \\
                0.2 &      0.336565 &       0.020660 \\
                0.4 &      0.491436 &       0.129221 \\
                0.5 &      0.485869 &       0.020968 \\
\bottomrule
\end{tabular}
\label{table:mse_spring}
\end{table}

\begin{table}[H]
\centering
\caption{MSE between the ground truth trajectory and model predicted trajectory for a given mass $m$. Each row corresponds to a column in Figure \ref{fig:spring_trj}(b), where $m$ is mapped to $m^*$ and evaluated while $k^*$ is held constant at the value corresponding to $k = 0.25$.} 
\begin{tabular}{lrrr}
\multicolumn{3}{c}{Mass-spring interpolated mass | MSE: t-span[0,20]} \\
\toprule
  $m$ &  Baseline MSE &  ConSciNet MSE \\
\midrule
 0.5 &      0.645808 &       0.353912 \\
 0.7 &      0.207255 &       0.010912 \\
 0.8 &      0.845841 &       0.163615 \\
 1.0 &      0.911760 &       0.525072 \\
\bottomrule
\end{tabular}
\label{table:mse_spring2}
\end{table}

In the ideal mass-spring trial the ConSciNet and Baseline models both returned a good disentangled latent representation with both models only encoding information in two of the provided latent variables. This was the expected behavior as only the mass and spring coefficient were varied in the input data. 

In Figure \ref{fig:spring_lv_dis} we present the disentanglement results for both models. The ConSciNet model encoded a spring coefficient-like value $k^*$ in the 1st latent variable position and a mass-like parameter $m^*$ in the 3rd latent variable position, Figure \ref{fig:spring_lv_dis}(a). The Baseline model encoded a spring coefficient-like value $k^*$ in the 2nd latent variable position and a mass-like parameter $m^*$ in the 3rd latent variable position, Figure \ref{fig:spring_lv_dis}(b). The remaining latent variable in both models presented as a flat surface with values close to 0 indicating little to no information was encoded. 

In the mass-spring trial we found the network parameterization learned in ConSciNet’s decoder maintained similar generative properties to the system’s underlying time evolution equations with regards to the physical parameters. As in the pendulum trial ConSciNet was able to generalize across the discovered parameter domain and extrapolate the time evolutions of the system governed by these parameters beyond the training interval t-span [0,10]. In Figure \ref{fig:spring_trj} we showcase trajectories generated from evaluating the decoders at four increasing values of $k^*$ and $m^*$.

The like parameter evaluation values were selected using a similar scheme to the pendulum trial to allow a ground truth comparison. Here instead of a cubic polynomial we fit a 5th order polynomial to map between the like parameters $k^*,m^*$ and their corresponding ground truth parameters $k, m$. A 5th order polynomial was selected to provided greater flexibility given the curvature present in Figure \ref{fig:spring_lv_dis}. This simple polynomial mapping between the like parameters and the ground truth parameter is sufficient to capture the physical behavior trend, however accuracy in both models is decreased in comparison to the pendulum trial, MSE Tables \ref{table:mse_pen}, \ref{table:mse_spring} and \ref{table:mse_spring2}. The need for establishing a robust mapping between the network disentangled parameters and some underlying physical parameter is a limitation of the current approach, particularly when the underlying physical parameters are unknown. In Figure \ref{fig:spring_trj}(a) and (b), we highlight the generalization ability of the models in regards to $k^*$ and $m^*$ respectively. As the spring coefficient like parameter $k^*$ and $m^*$ are increased independently the amplitude of the momentum $p$ increases aligning with the expected physical behavior. 

\section{Conclusion}

We have presented a neural framework for joint parameter discovery and generative modeling of Hamiltonian systems. Our approach merges aspects of deep latent variable models and Hamiltonian Neural Networks. We evaluated our approach on noisy simple dynamical systems. Here we demonstrated that it discovered parameters akin to the underlying physical parameters and returned a Hamiltonian parameterization that generalized well with respect to the disentangled parameters. Our approach respected the conservation of energy constraint and outperformed a non-physically constrained baseline model. The baseline model struggled to extrapolate beyond the training interval and lost/gained energy. The conservation of energy constraint embedded in our model limits our approach to energy conserving systems. The decoder and auxiliary latent variables used in our approach could be modified to embed another constraint. For example, the decoder could be replaced with an even or odd network to embed an even or odd symmetry. We believe that merging physical constraint embedded neural networks and deep latent variable models can provide a strong framework for improving network interpretability and generalization ability. 

\bibliographystyle{unsrt}

\bibliography{refs}

\end{document}